\documentclass{article}

\usepackage{PRIMEarxiv}

\usepackage[utf8]{inputenc} 
\usepackage[T1]{fontenc}    
\usepackage{hyperref}       
\usepackage{url}            
\usepackage{booktabs}       
\usepackage{amsfonts}       
\usepackage{nicefrac}       
\usepackage{microtype}      
\usepackage{lipsum}
\usepackage{fancyhdr}       
\usepackage{graphicx}       
\graphicspath{{media/}}     

\title{Many-MobileNet:  Multi-Model Augmentation for Robust Retinal Disease Classification  
}

\usepackage[affil-it]{authblk}  

\author[2]{Hao Wang $^{1,}$\thanks{Corresponding author} , Wenhui Zhu} 
\author[2]{Xuanzhao Dong} 
\author[2]{Yanxi Chen} 
\author[2]{Xin Li} 
\author[3]{Peijie Qiu} 
\author[1]{Xiwen Chen}
\author[2]{Vamsi Krishna Vasa}
\author[2]{Yujian Xiong}
\author[4]{Oana M. Dumitrascu}
\author[1]{Abolfazl Razi} 
\author[2]{Yalin Wang}

\affil[1]{School of Computing, Clemson University, SC, USA} 
\affil[2]{School of Computing and Augmented Intelligence, Arizona State University, AZ, USA}
\affil[3]{McKeley School of Engineering, Washington University in St. Louis, MO, USA}
\affil[4]{Department of Neurology, Mayo Clinic, AZ, USA}

\begin{document}

\maketitle

\begin{abstract}
In this work, we propose Many-MobileNet, an efficient model fusion strategy for retinal disease classification using lightweight CNN architecture. Our method addresses key challenges such as overfitting and limited dataset variability by training multiple models with distinct data augmentation strategies and different model complexities. Through this fusion technique, we achieved robust generalization in data-scarce domains while balancing computational efficiency with feature extraction capabilities. 
Our software package is available at  \href{https://github.com/Retinal-Research/Beyond-MobileNet}{https://github.com/Retinal-Research/NN-MOBILENET} 

\end{abstract}


 \section{Introduction}

Retinal diseases (RD) are among the leading causes of visual impairment and blindness worldwide, particularly in cases of myopic maculopathy, which pose significant challenges in clinical diagnosis \cite{retinaldisease,deeplearning4,rdmeterial,wang2024rbad}. Specifically, diabetic retinopathy (DR) is a major contributor to blindness among working-aged adults globally \cite{DLI,zhu2023nnmobile}. Early detection and accurate grading of DR are critical for timely intervention, which can prevent severe vision loss \cite{vbxkckml,greenchannel, wang2022fast}. In clinical practice, grading DR based on retinal images is crucial for determining the progression of the disease and guiding treatment decisions \cite{green,comp-expertab,deeplearning3,deeplearning5,likassa2024robust}. 

Deep learning-based automated diagnostic tools have demonstrated significant potential in assisting clinicians with RD detection and monitoring \cite{dai2021deep,liu2022deepdrid,vasa2024context}. Over the years, convolutional neural networks (CNNs) \cite{CANet,zoom-in-net,sem+adv,AFN,comp-expertab,DETACH,selfmil} and, more recently, Vision Transformers (ViTs) \cite{khan2022transformers} have emerged as the primary techniques in medical image analysis due to their ability to extract and analyze critical features from retinal images \cite{zhu2023otre,zhu2023optimal}.
While Vision Transformers have gained popularity due to their ability to capture long-range dependencies, they often require large datasets and come with increased model complexity, making them prone to overfitting, particularly in medical image tasks where data is scarce. On the other hand, CNNs, with their simpler architectures, remain highly effective for tasks like retinal disease classification, where localized feature extraction is crucial for accurate diagnosis. For these reasons, CNN-based architectures continue to be widely used for tasks that demand both efficiency and accuracy \cite{seg,seg2,dai2021deep,liu2022deepdrid}.

In this work, we apply nnMobileNet \cite{zhu2023nnmobile}, a lightweight CNN architecture, to the retinal image quality classification task \cite{sandler2018mobilenetv2}. To address challenges such as overfitting and limited dataset variability \cite{dropout_chanelwise,channelattention,ReXNet}, we propose a model fusion strategy. This strategy combines multiple lightweight nnMobileNet models that use the same architecture but with different levels of model complexity, each trained with different data augmentation techniques. By fusing these models, we improve generalization and ensure that the final predictions are robust under a wide range of conditions, even in data-scarce environments.

 \section{Methods}
\label{method}

To optimize the performance of nnMobileNet for the retinal image quality classification task, we employed a comprehensive approach involving multiple hyperparameter adjustments and model enhancements. Additionally, we explored model width scaling and conducted extensive tests. Combined with a model fusion strategy, these refinements were essential in balancing computational efficiency with robust feature extraction. 
Below, we discuss each modification and its influence on the model's overall performance and generalization ability.

\begin{figure}[]
    \centering
    \includegraphics[width=1\textwidth]{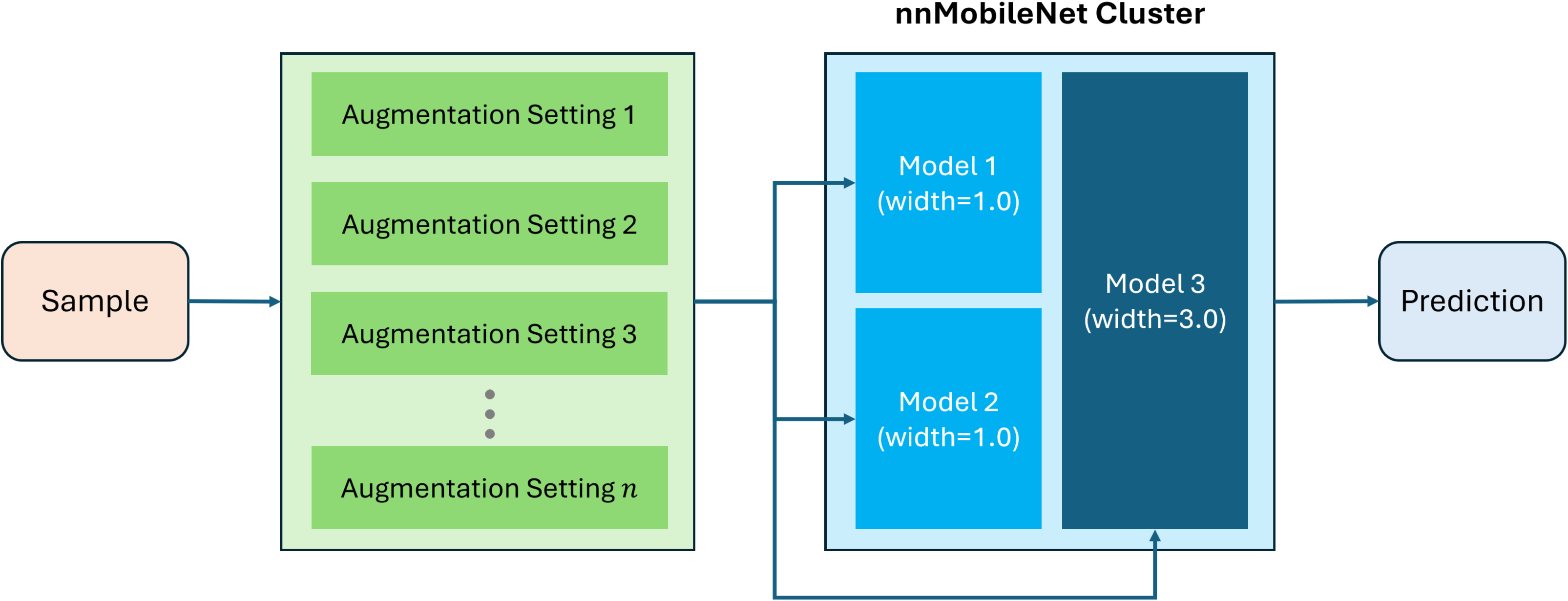}

    \caption{Concept of model fusion during inference.}
    \label{fig:frame}

\end{figure}

\subsection{Model Architecture }
In this work, we utilize nnMobileNet to balance the trade-offs between model efficiency and accuracy while maintaining robustness against overfitting, particularly when handling small and imbalanced medical datasets \cite{sandler2018mobilenetv2}. Due to its lightweight CNN architecture, nnMobileNet features depthwise separable convolutions combined with linear bottleneck layers, which minimize computational costs while retaining high representational capacity \cite{zhu2023nnmobile}. 
Key improvements in our version of nnMobileNet include the integration of advanced channel-wise attention mechanisms \cite{channelattention}, especially Squeeze-and-Excitation (SE) blocks \cite{hu2018squeeze}, which recalibrate feature maps and improve the focus on relevant retinal features. This design enables our model to efficiently process high-resolution fundus images without the need for excessively deep or complex networks. 
Specifically, the number of feature layers can be controlled by the channel multiplication factor. 
In this work, we compose a simple and efficient architecture $-$ Many-MobileNet $-$ that fuses multiple nnMobileNet trained on the same dataset with different model settings to enhance the robustness during inference on new samples, as shown in Figure \ref{fig:frame}.

\subsection{Data Augmentation Strategy }
Given our dataset's limited quantity and uniform nature, training a high-performance model solely on the existing data is insufficient. The dataset's lack of diversity makes it difficult for a single model to generalize well across various cases. 
By applying different augmentation techniques to different models \cite{LAT,SatFormer-B}, we enhance the training data and improve the overall robustness and generalization of the combined model during the inference stage. 

The key difference in the data augmentation strategy lies in using different normalization techniques. Models trained on general datasets such as ImageNet utilize customized normalization values \cite{liu2022convnet}, whereas models trained on medical image classification tasks employ a different set of normalization parameters. 
These variations in normalization result in differences in model performance during both training and inference, as the models were tuned to their respective datasets.

\subsection{Training Strategy}
Previous studies \cite{LAT,SatFormer-B,CANet,MIL-VT} have suggested that excessive data augmentation may compromise the integrity of fundus images, which led to the use of limited augmentations such as spatial transformations and brightness adjustments in retinal fundus image tasks. However, based on our empirical findings, these basic augmentations were insufficient to eliminate overfitting in RD tasks. To address this, we conducted exploratory experiments to test various data augmentation combinations that could prevent overfitting and improve model robustness.

The training strategy in this experiment involved the systematic exploration of several hyperparameters to optimize model performance. We experimented with different normalization settings, and dropout rates ranging from $0$ to $0.10$ were tested to regularize the models and prevent overfitting. Additionally, batch sizes of $8$, $16$, and $32$ were evaluated to assess their effects on model convergence and generalization. The learning rates tested included $1e-3$, $1e-4$, $1e-5$, and $1e-6$, each decayed using a cosine learning rate scheduler to ensure smooth convergence throughout the training process.
We also explored different model widths by adjusting the width multiplier, comparing lightweight models with a channel multiplier of $1.0$ to medium-sized models with a channel multiplier of $3.0$. This allowed us to balance between model complexity and performance, with the lightweight models focused on computational efficiency, while the medium-sized models captured more complex features. A detailed training setting is given in Table \ref{tb:config}. Through this comprehensive exploration of hyperparameters, we were able to identify the optimal configurations that enhanced model performance in this retinal image classification task.

\subsection{Implementation Details}
In this experiment, we utilized multiple \texttt{NVIDIA V100} GPUs for training. All models were trained using the PyTorch framework, with standardized input image sizes of $224 \times 224$. 

\begin{table}[]
    \centering
    \caption{All training configurations for Many-MobileNet.}
    \begin{tabular}{c|c}
        \toprule
         \multicolumn{2}{c}{Training configuration} \\
        \midrule
         Optimizer & AdamP \\
         Batch size & 8, 16, 32\\
         Learning rate & 1e-3, 1e-4, 1e-5, 1e-6\\
         Weight decay & 0.005 \\
         Scheduler & Cosine decay \\
         Dropout rate & 0.01, 0.02, 0.05\\
         Epoch & 500\\
         Loss & Cross-entropy \\
         Metric & acc, auc, average\\
         Model width & 1.0, 3.0\\
        \bottomrule
    \end{tabular}

    \label{tb:config}
\end{table}

As illustrated in Table \ref{tb:config}, a weight decay of $0.005$ was applied throughout the training to prevent overfitting, particularly in the deeper layers of the network. The models were trained for $500$ epochs to ensure full convergence.
The \texttt{AdamP} optimizer was employed due to its effectiveness in enhancing generalization and stabilizing the training process \cite{adamp}. In addition, dropout rates between $0.00$ and $0.10$ were applied to regularize the models and prevent overfitting. Model widths (channel multiplier) of $1.0$ (lightweight) and $3.0$ (medium-sized) were tested to evaluate the trade-off between model capacity and computational efficiency.
Throughout the training process, the models were evaluated using key metrics, including accuracy (acc) to measure the percentage of correctly classified instances, the area under the curve (AUC) to assess the model’s ability to distinguish between classes, and an average metric to provide a comprehensive evaluation of the model’s overall performance. The cross-entropy loss function was applied across all models to ensure robust performance in the classification tasks.

\subsection{Model Fusion }
In our approach, model fusion plays a critical role in overcoming the limitations of the dataset’s uniformity and small sample size. The fusion of multiple models, each trained under different conditions and with distinct data augmentation strategies, is the key to enhancing the model’s robustness and overall performance. This fusion technique ensures that each model contributes complementary strengths, leading to more accurate and reliable predictions.

\begin{figure}[]
    \centering
    \includegraphics[width=1\textwidth]{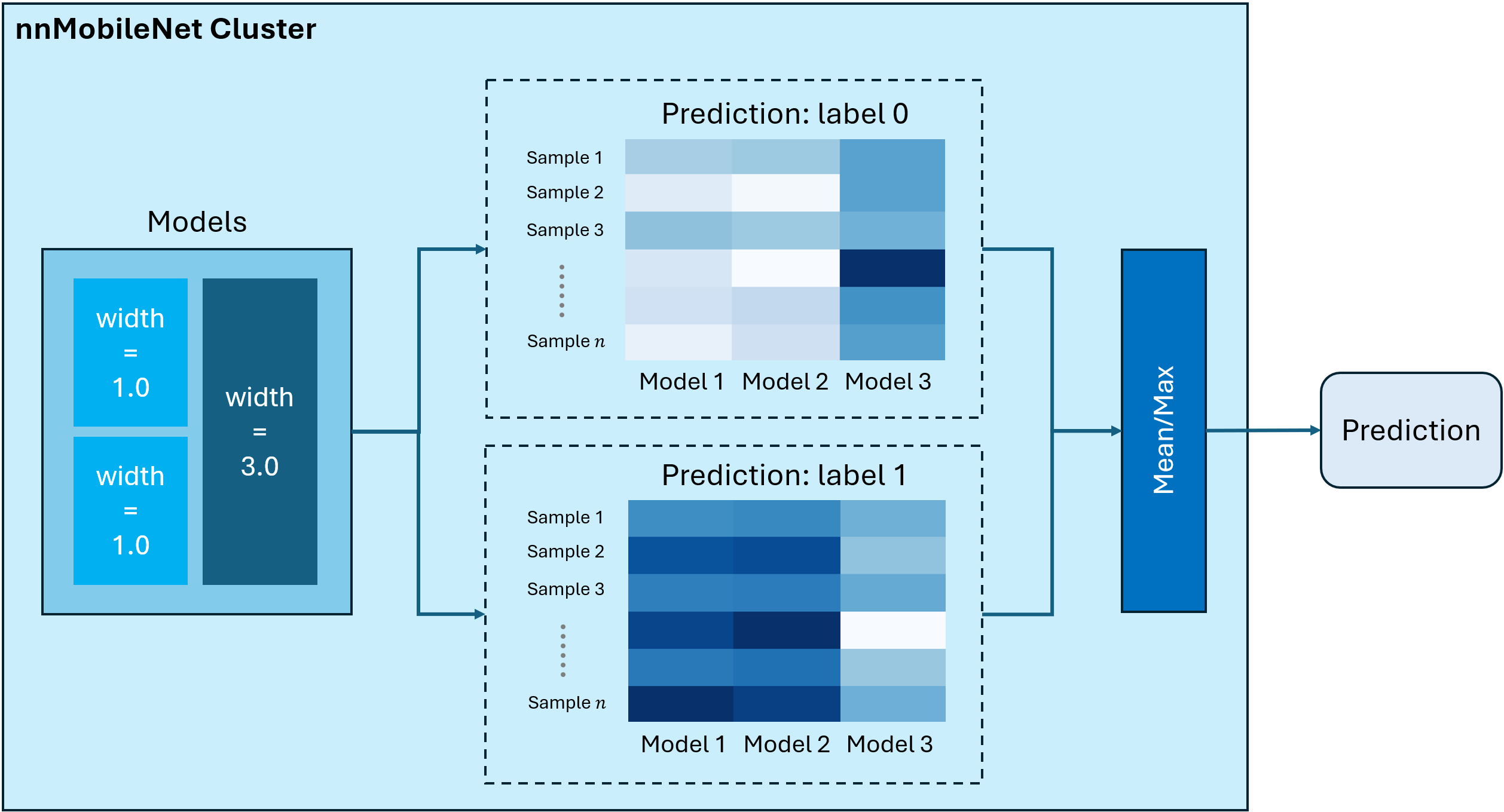}

    \caption{Prediction of Many-MobileNet.}
    \label{fig:pred}

\end{figure}

In this work, the model width (channel multiplier) plays a critical role in determining the network's capacity and resource requirements. The model width scales the number of channels in each layer, which significantly impacts the model's parameters and size. 
For instance, with a model width equal to $1.0$, the model maintains the base number of channels (e.g., $32$, $64$, $128$, etc.), resulting in a smaller model approximately $13MB$ in size. However, when a model width is $3.0$, the channels in each layer are tripled, causing a significant increase in parameters and resulting in a model size of around $120MB$. 
This makes the model width a key factor in balancing between computational efficiency and the ability to capture complex features.

Our model fusion involves the combination of multiple nnMobileNet models, each deployed with different channel multipliers. We employ two lightweight models with a channel multiplier of $1$, optimized for efficiency, and one medium-sized model with a channel multiplier of $3$, designed for more complex feature extraction. This architectural diversity ensures that the lightweight models contribute speed and computational efficiency, while the larger model provides the capacity to capture more intricate details in the retinal images.
Each model is trained with a distinct data augmentation strategy to further enhance the diversity of learned features, and ensure that each model learns a unique representation of the retinal features, capturing different aspects of the image. During inference, this fusion reduces the risk of overfitting to specific augmentations or data patterns.

During inference, we employ a prediction voting method to combine the predictions from each model. The final decision is calculated by the maximum or average outputs of the two lightweight models and the medium-sized model, while some models may carry different results that are more certain than others, as shown in Figure \ref{fig:pred}. This method provides redundancy, ensuring that the final prediction benefits from the strengths of all models.

\section{UWF4DR - Quality assessment for ultra-widefield fundus images}

\subsection{Dataset and Evaluation Metrics}

\begin{figure}[]
    \centering
    \includegraphics[width=1\textwidth]{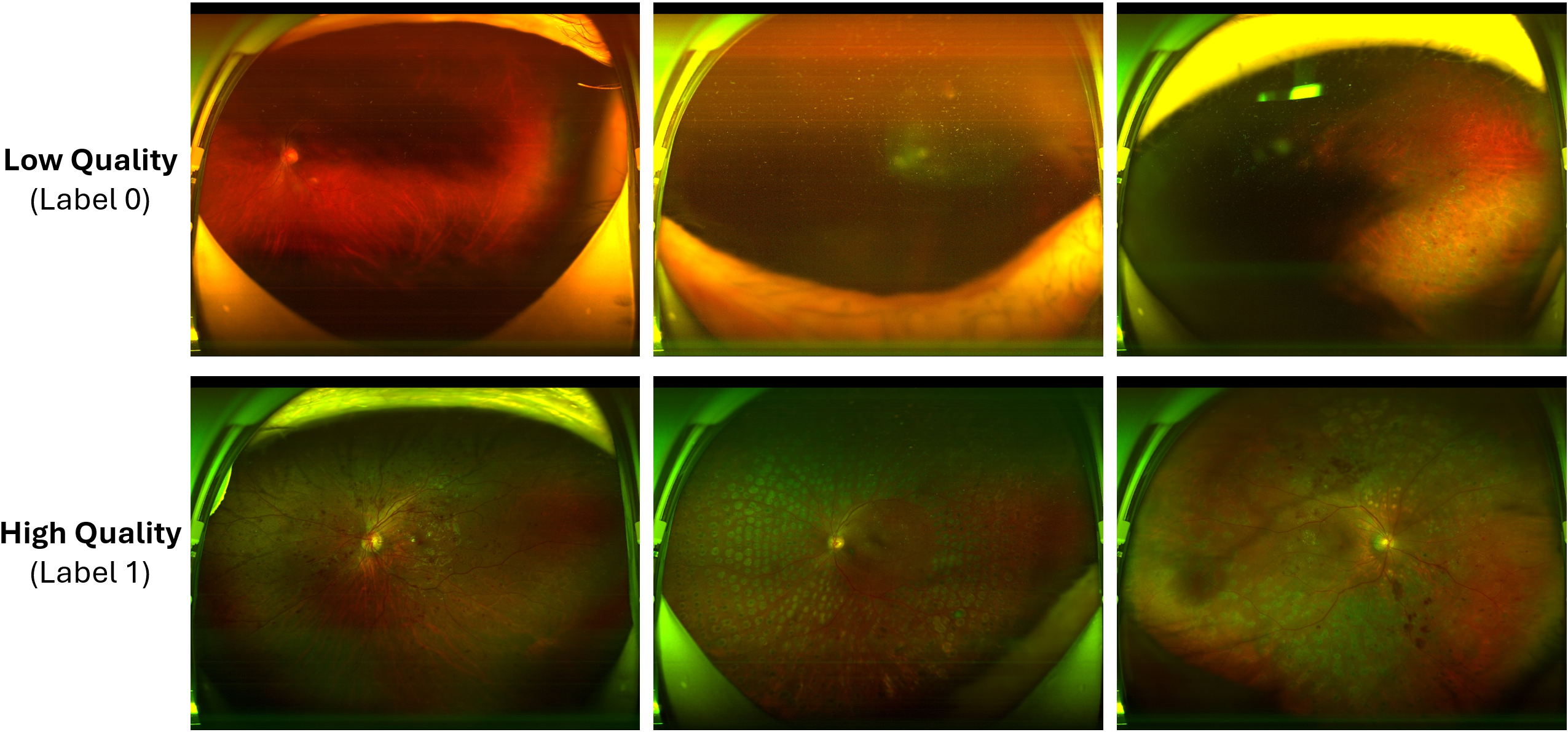}

    \caption{Sample images of UWF4DR dataset, where label 0 represents the ungradable images and label 1 represents gradable images.}
    \label{fig:sample}

\end{figure}

The images used in this study comes from the ultra-widefield (UWF) fundus imaging for diabetic retinopathy (DR) dataset, which aims to advance automatic DR analysis from UWF fundus images. The dataset includes UWF images with up to a 200-degree view of the retina, allowing the identification of predominantly peripheral lesions (PPL) that are present in a significant portion of eyes with DR. The images are classified into different DR stages based on the International Clinical Diabetic Retinopathy (ICDR) Severity Scale, ranging from no apparent retinopathy to proliferative diabetic retinopathy (PDR), including diabetic macular edema (DME). The dataset is divided into three tasks: image quality assessment, DR classification, and DME classification. 

This study specifically focuses on image quality assessment for ultra-widefield fundus images. The dataset of this task contains a total of $434$ samples that include $205$ ungradable samples and $229$ gradable samples, their differences are illustrated in Figure \ref{fig:sample}.
This comprehensive dataset provides a foundation for developing algorithms that assist in the timely diagnosis and management of DR patients, particularly by reducing the manual effort required for grading UWF fundus images.

\subsection{Experimental Results}

In this experiment, we conducted a comprehensive evaluation by submitting multiple models, each trained with different parameter configurations. 
We investigated several models to analyze the impact of different hyperparameter settings. The models were also evaluated using three key metrics: AUC, accuracy, and an average metric.
\begin{figure}[]
    \centering
    \includegraphics[width=1\textwidth]{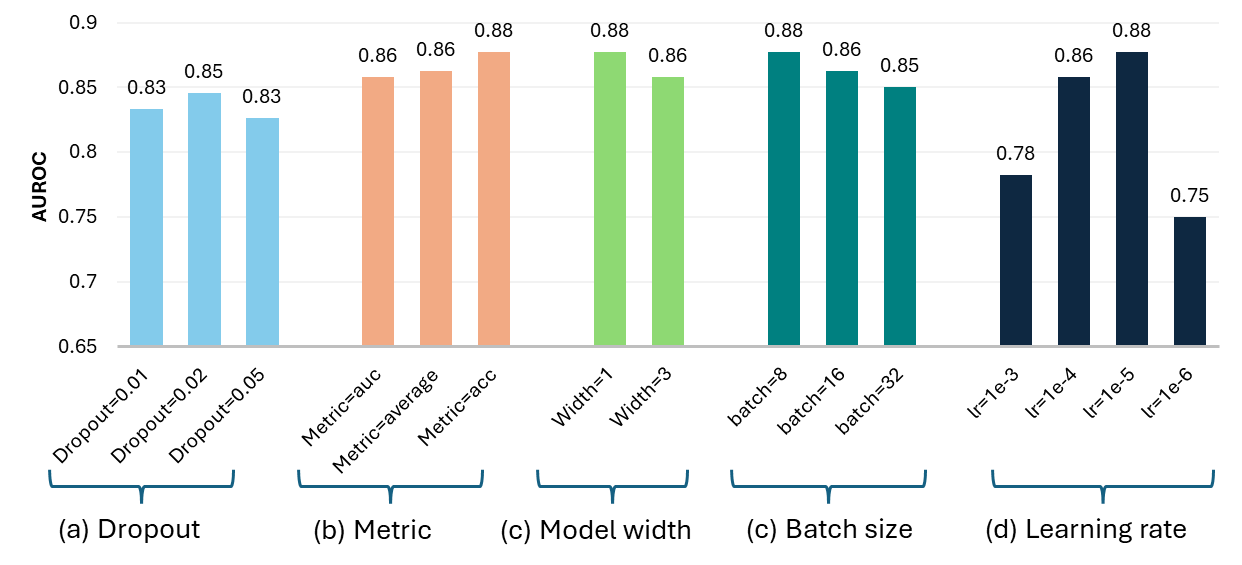}

    \caption{Parameter empirical studies based on UWF4DR dataset.}
    \label{fig:bar}

\end{figure}
As shown in Figure \ref{fig:bar}, we tested two channel multiplier settings, where the results suggest that the lightweight model provided better results on the validation set. Also, a batch size of 8 achieved the highest accuracy, while larger batch sizes of 16 and 32 resulted in slightly lower performances. Finally, we tested learning rates of 1e-3, 1e-4, 1e-5, and 1e-6. The learning rate of 1e-5 produced the best accuracy, followed by 1e-4. Higher learning rates, such as 1e-3 and a lower learning rate of 1e-6, performed worse, respectively.

From these models, we selected the top three models based on their performance on the validation set, focusing primarily on the AUROC (Area Under the Receiver Operating Characteristic Curve) metric. These top-performing models were then analyzed further to understand their strengths and weaknesses.
\begin{table}[]
    \centering
    \caption{Best model weights performed at the official validation set.}
    \resizebox{1\textwidth}{!}{
    \begin{tabular}{c|ccccccccc}
        \toprule
         Method   & AUROC 
& AUPRC 
& Sensitivity 
& Specificity 
& metric
& lr
& width& batch
&dropout\\ \midrule
Weight1 & 0.8772
&  0.9069
& 0.9729
& 0.7083
& acc
& 1e-5
& 1.0& 8&0.01
\\
Weight2 & 0.8626
&  0.8676
& 0.7567
& 0.8333
& average
& 1e-5
& 1.0& 16&0.02\\  
         Weight3 & 0.8581 
&  0.8680& 0.7568
& 0.875
& auc
& 1e-4
& 3.0& 16&0.05 
\\ \hline
        Ensemble with max & 0.8468&  0.8570& 0.8648& 0.75& - & - & - & - & -\\
         Ensemble with average & 0.8125&  0.8043& 1.0& 0.625& - & - & - & -  & -\\
        \bottomrule
    \end{tabular}
    }
    \label{tb:val}
\end{table}
As shown in Table \ref{tb:val}, models with lower learning rates exhibited more stable performance. 
Secondly, batch size also affected model outcomes. Smaller batches (e.g., 8) improved sensitivity, helping the model better identify positive samples, while larger batches enhanced specificity, improving the model’s ability to classify negative samples.
Meanwhile, lower dropout values helped maintain the model's generalization ability, while higher dropout values led to information loss, negatively impacting performance on the validation dataset. However, it was crucial for preventing overfitting on the actual testing dataset. 
However, the fusion of the models did not perform well in the validation dataset. This performance decrement might be caused by the small and potentially unrepresentative validation data. 


\begin{table}[]
    \centering
    \caption{Final ranking based on official testing set.}
\begin{tabular}{c|cccccc}
        \toprule
         Team   & AUROC
& AUPRC
&  Sensitivity
&Specificity  &Time
 &Ranking score
\\
        \midrule
         Rank 1st & 0.9716&  0.9820&  0.9322& 0.950& 0.0732&0.9679\\
         Rank 2nd & 0.9622&  0.9759&  0.8305& 0.975& 0.0573&0.9594\\   \hline  
         Ours (Rank 3rd) & 0.9525&  0.9683&  0.8983& 0.925
& 0.1098&0.9470\\   

        \bottomrule
    \end{tabular}

    \label{tb:rank}
\end{table}
In the final ranking, we submitted the combined model of our selected top 3 models. As shown in Table \ref{tb:rank}, our team's model fusion strategy secured 3rd place with specific metrics of AUROC 0.9525 and AUPRC 0.9683, demonstrating robust classification performance. The Sensitivity 0.8983 and Specificity 0.925 indicate a balanced ability to identify both positive and negative samples. Although the model's computation time was longer at 0.1098 due to multiple inference times, our model fusion strategy successfully combined the strengths of lightweight and medium-sized models, enhancing computational efficiency while extracting complex features. Ultimately, the fusion significantly improved the model's generalization.

\section{Conclusion}
In conclusion, we introduced Many-MobileNet $-$ an architecture consisting of multiple nnMobileNet for retinal image quality classification. By combining lightweight and medium-sized models, each trained with different data augmentation techniques, our approach effectively mitigates overfitting and improves model robustness. The fusion strategy takes advantage of the computational efficiency of smaller models and the feature extraction power of more complex models, resulting in a balanced system that excels in accuracy and generalization. The final result 
 of this method
demonstrates its possibility for broader applications in medical image analysis and its potential to generalize well in data-limited challenges.

\bibliographystyle{unsrt}  
\bibliography{refs}

\end{document}